# A SURVEY OF ARABIC DIALOGUES UNDERSTANDING FOR SPONTANEOUS DIALOGUES AND INSTANT MESSAGE

AbdelRahim A. Elmadany[1], Sherif M. Abdou[2] and Mervat Gheith[1]

[1] Institute of Statistical Studies and Research (ISSR), Cairo University
[2] Faculty of Computers and Information, Cairo University

## ABSTRACT

*Building dialogues systems interaction has recently gained considerable attention, but most of the resources and systems built so far are tailored to English and other Indo-European languages. The need for designing systems for other languages is increasing such as Arabic language. For this reasons, there are more interest for Arabic dialogue acts classification task because it a key player in Arabic language understanding to building this systems. This paper surveys different techniques for dialogue acts classification for Arabic. We describe the main existing techniques for utterances segmentations and classification, annotation schemas, and test corpora for Arabic dialogues understanding that have introduced in the literature*

## 1.INTRODUCTION

Build a completely Human-Computer systems and the belief that will happens has long been a favourite subject in research science. Consequently, dialogues language understanding is growing and considering the important issues today for facilitate the process of dialogue acts classification. Human-Computer system typically consist of the main components as shown in Figure 1 (Lee et al.,2010).

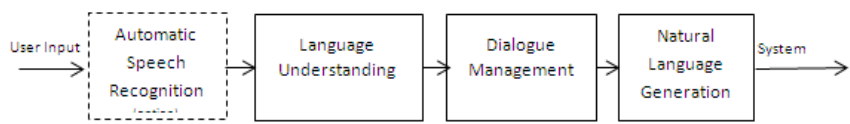

Figure1. Traditional Architecture of Dialog System

**User Input:** User input is usually speech signal with noises in spoken dialogue system or textual input in chat.

- **Automatic Speech Recognition (ASR):** it is using only on spoken dialogue system and not use in written 'chat' dialogue system. Therefore, this component is an option and it is converting a sequence of parameter vectors in waveform from speech signal processing into a textual input.

- **Language Understanding (LU):** Analyse the input textual "turn" by natural language processing (NLP) tools (e.g., morphological analysis, part-of-speech tagging, and shallow parsing). The LU maps the utterances to a meaning representation or semantic representation e.g. dialog act, user goal, and named entities; Therefore, this component what we interested here.

- **Dialogue Management (DM):** it is considering the core of dialogue system because it coordinates the activity of all components, controls the dialog flow, and communicates with external applications. The DM plays many roles, which include discourse analysis, knowledge database query, and system action prediction based on the discourse context.

- **Natural Language Generation (NLG):** The system responses are typically generating as natural language with a list of content items from a part of the external knowledge database (e.g., bank customer-service database) that answers the specific user query or request.

- **System Output:** the dialogue system export two different outputs based on its type; first, a text when use a written dialogues and second, a speech signal by text-to-speech (TTS) tools when use a spoken dialogue.

Arabic is one of a class of languages where the intended pronunciation of a written word cannot be completely determining by its standard orthographic representation; rather, a set of special diacritics are needs to indicate the intended pronunciation. Different diacritics for the same spelling form produce different words with maybe different meanings. These diacritics, however, are typically omitted in most genres of written Arabic, resulting in widespread ambiguities in pronunciation and (in some cases) meaning. While native speakers are able to disambiguate the intended meaning and pronunciation from the surrounding context with minimal difficulty, automatic processing of Arabic is often hampered by the lack of diacritics. Text-to-speech (TTS), Part-Of-Speech (POS) tagging, Word Sense Disambiguation, and Machine Translation (ML) can be enumerated among a longer list of applications that vitally benefit from automatic discretization(Al-Badrashiny,2009). Moreover, there are three categories of Arabic language: Classic Arabic "The old written form", Modern Standard Arabic (MSA) "The famous written form today", and dialectal Arabic "Native spoken languages of Arabic speakers" (Diab and Habash,2007).. Since, the written form of the Arabic language - MSA- is differs from dialectal Arabic.  However, MSA used primarily for written form but the regional dialects is prevalence in spoken communications or day-to-day dealings. Unlike MSA, the dialects does not have a set of written grammars rules and have different characteristics e.g. morphology, syntax and phonetics.

Moreover, Dialectal Arabic can mainly divided into six dialects groups: Maghrebi, Egyptian, Levantine, Gulf, Iraqi and other. Those regional dialects of Arabic are differ quite a bit from each other.  Egyptian dialect commonly known as Egyptian colloquial language is the most widely understood Arabic dialect (Zaidan and Callison-Burch, 2012).

In this paper, we focus on language understanding component for Arabic dialogues system. However, there are few works have developed for Arabic spoken dialogue system either MSA or dialect as the best of our knowledge; this is mainly due to the lack of tools and resources that are necessary for the development of such systems (Zaghouani, 2014; Lhioui et al., 2013). Therefore, building language-understanding component for dialogue system is requiring four parts: (1)

Dialogue Acts Annotation Schema (2) Dialogue corpus (3) Segmentation Classification (4) Dialogue Acts Classification; consequently, this paper present a survey for these parts.

This paper is organized as follows: section 2 present the concepts and terminology that's used in the paper, section 3 present Arabic language understanding components (dialogue acts annotation schema, dialogue corpus, segmentation classification, and dialogue acts classification); and finally the conclusion and feature works are reported in section 4.

## 2. CONCEPTS AND TERMINOLOGIES

This section present the concepts that related to language understanding and used in this paper.

### 2.1. Dialogue Act

The terminology of speech acts has been addressed by Searle (1969) based on Austin work (1962) as (Webb, 2010):

— Assertive commit the speaker to the truth of some proposition (e.g. stating, claiming, reporting, announcing)
— Directives attempts to bring about some effect through the action of the Hearer (e.g. ordering, requesting, demanding, begging)
— Commissures commit speaker to some future action (e.g. promising, offering, swearing to do something)
— Expressive are the expression of some psychological state (e.g. thanking, apologizing, congratulating)
— Declarations are speech acts whose successful performance brings about the correspondence between the propositional content and reality (e.g. resigning, sentencing, dismissing, and christening).

Dialogue act is approximately the equivalent of the speech act of Searle (1969). Dialog acts are different in different dialog systems. So, Major dialogue theories treat dialogue acts (DAs) as a central notion, the conceptual granularity of the dialogue act labels used varies considerably among alternative analyses, depending on the application or domain(Webb and Hardy, 2005). Hence, within the field of computational linguistics - recent work - closely linked to the development and deployment of spoken language dialogue systems, has focused on the some of the more conversational roles such acts can perform. Dialogue act (DA) recognition is an important component of most spoken language systems. A dialog act is a specialized speech act. DAs are different in different dialog systems. The research on DAs has increased since 1999, after spoken dialog systems became commercial reality (Stolcke et al., 2000). So, (Webb, 2010) define the DAs as the labelling task of dialogue utterance that serve in short words a speaker's intention in producing a particular utterance.

### 2.2. Turn vs Utterance

In natural human conversation, turn refer to the speaker talking time and turn-taking refer to the skill of knowing when we start and finish the turn in the conversion. The turn boundary contains one or more sentences moreover, the "turn-taking" is generally fixed to the expression of a single sentences. In the spoken dialogue system the term of utterance is refer to the one speech act. (Traum and Heeman, 1997) has defines the utterance unit by one or more of the following factors:

1. Speech by a single speaker, speaking without interruption by speech of the other, constituting a single Turn.
2. Has syntactic and/or semantic completion.
3. Defines a single speech act.
4. Is an intonational phrase.
5. Separated by a pause.

Consequently, this paper refers to an utterance as a small unit of speech that corresponds to a single act(Webb, 2010; Traum and Heeman, 1997). In speech research community, utterance definition is a slightly different; it refers to a complete unit of speech bounded by the speaker's silence while, we refer to the complete unit of speech as a turn. Thus, a single turn can be composed of many utterances. Moreover, turn and utterance can be the same definition when the turn contains one utterance as used in(Graja et al., 2013). Here an example of a long user *turn* from Arabic dialogues corpus that contains many utterances (Elmadany et al., 2014):

| Arabic | كنت عايزة افتح دفتر توفير عايزة اسأل على الإجراءات كنت عايزة اسألك لو سمحت |
|---|---|
| Buckwalter | lw smHt knt EAyzp As>lk knt EAyzp AftH dftr twfyr EAyzp As>l Ely Al<jrA'At |
| English | Excuse me I want to ask you I want open an account I need to know the proceeds |

This *turn* contains four *utterances* as:

1. [ لو سمحت ] [ *lw smHt* ][ excuse me]
2. [ كنت عايزة اسألك ] [ *knt EAyzp As>lk* ] [I want to ask you]
3. [ كنت عايزة افتح دفتر توفير ] [ *knt EAyzp AftH dftr twfyr* ] [I want open an account]
4. [عايزة اسأل على الإجراءات] [*EAyzp As>l Ely Al<jrA'At* ] [ I need to know the proceeds]

## 2.3 Dialectal Arabic

There are three categories of Arabic language: Classic Arabic "The old written form", Modern Standard Arabic (MSA) "The famous written form today", and dialectal Arabic "Native spoken languages of Arabic speakers" (Diab and Habash, 2007).Moreover, the written form of the Arabic language - MSA- is completely differs from dialectal Arabic. However, MSA used primarily for written form but the regional dialects is prevalence in spoken communications or day-to-day dealings. Unlike MSA, the dialects does not have a set of written grammars rules and have different characteristics e.g. morphology, syntax and phonetics.

Dialectal Arabic can mainly divided into six dialects groups: Maghrebi, Egyptian, Levantine, Gulf, Iraqi and other. Those regional dialects of Arabic are differ quite a bit from each other.

Egyptian dialect commonly known as Egyptian colloquial language is the most widely understood Arabic dialect (Zaidan and Callison-Burch, 2012).

## 3. LANGUAGE UNDERSTANDING COMPONENT

In this section, we present the recent researches for the four parts of building language-understanding component for Arabic dialogue systems, these parts are (1) Dialogue Acts Annotation Schema (2) Dialogue corpus (3) Segmentation Classification (4) Dialogue Acts Classification.

## 3.1. Dialogue Acts Annotation Schema

The idea of dialogue act plays a key role in studies of dialogue, especially in communicative behaviour understanding of dialogue participants, in building annotated dialogue corpora and in the design of dialogue management systems for spoken human-computer dialogue. Consequently, to build annotated dialogues corpus we need annotation schema that contains a list of predefined categories, semantic labels, or dialogue acts; schema is considering the key player to build the annotated corpus and dialogue acts classification task.

Searle (1969) has addressed the history of dialogue acts schema (see section 2.1). Moreover, the research on dialogue acts is increasing since 1999 after spoken dialogue systems become a commercial(Stolcke et al., 2000). Many dialogue acts schema applied in non-Arabic dialogues such as English and Germany; below we present most of them:

— The MapTask project (Anderson et al., 1991) proposed labelling schema using 12 dialogue acts based on two categories initiating moves and response:

- **Initiating moves includes**
  - Instruct
  - Explain
  - Check
  - Align
  - Query-yn
  - Query-w
- **Response moves includes**
  - Acknowledge
  - Reply-y
  - Reply-n
  - Reply-w
  - Clarify
  - Ready

— The VERBMOBIL project (1993-2000) aimed at the development of an automatic speech to speech translation system for the languages German, American English and Japanese (Wahlster, 2000).Moreover, the VERBMOBIL Project had two phases, the first phase proposed labelling schema using hierarchy of 43 dialogue acts(Jekat et al., 1995) as shown in Figure 2; the second phase expanded the dialogues from meeting scheduling to comprehensive travel planning; thus change labelling schema to a hierarchy of 18 dialogue acts(Alexandersson et al., 1998):

- Thank
- Deliberate
- Bye
- Request-suggest
- Greet
- Request-comment
- Suggest
- Accept
- Reject
- Init
- Digress
- Clarify
- Give-reason
- Motivate
- Garbage
- Feedback
- Confirm
- Introduce

— The DAMSL (Dialogue Act Markup using Several Layers) has proposed as a general-purpose schema (Allen and Core, 1997; Core and Allen, 1997; Core et al., 1998) developed for multi-dimensional dialogue acts annotation. Moreover, Jurafsky et al. (1997) reported an improved version of DAMSL to annotate a large amount of transcribed speech data 'Switchboard Corpus' because of the difficulty of consistently applying the DAMSL annotation schema(Jurafsky et al., 1998; Jurafsky et al., 1997). Consequently, SWITCHBOARD-DAMSL schema includes 220 dialogues acts, but it is still difficult to be used for manual annotation because it is a very large set, and Jurafsky et al. (1997) reported 0.80 of Kappa score with the 220 dialogue acts and 130 dialogue acts occurred less than 10 times in the entire corpus(Webb, 2010). Therefore, to obtain enough data per class for statistical modelling purposes, Jurafsky et al. (1997) proposed new dialogue act schema namely SWITCHBOARD contains 42 mutually exclusive dialogue acts types:

- Statement-non-opinion
- Collaborative Completion
- Acknowledge
- Repeat-phrase
- Statement-opinion
- Open-question 0.3%
- Abandoned/Uninterpretable
- Rhetorical-questions
- Agree/accept
- Hold before answer
- Appreciation
- Reject
- Yes-no-question
- Negative non-no answers
- Non-verbal
- Signal-non-understanding
- Yes answers
- Other answers
- Conventional-Closing
- Conventional-opening
- Wh-question
- Or-clause
- No answers
- Disprefered answers
- Response acknowledgement
- 3rd-party-talk
- Hedge
- Offers, options commits
- Declarative yes-no-question
- Self-talk
- Other
- down-player
- Back-channel in question form
- Maybe/accept-part
- Quotation
- Tag-question
- Summarise/Reformulate
- Declarative wh-question
- Affirmative non-yes answers
- Thanking
- Action-directive
- Apology

— The ICSI-MRDA Meeting Room corpus used a variant of the DAMSL dialogue acts schema like the SWITCHBOARD corpus by combining the tags into single, distinct dialogue acts to reduce aspects of the multidimensional nature of the original DAMSL annotation scheme. There are 11 general tags and 39 specific acts that are used over ICSI-MRDA Meeting Room corpus (Shriberg et al., 2004). So, the AMI project, a European research project centred on multi-modal meeting room technology, uses 15 dialogue acts:

- Back-channel
- Assess
- Stall
- Elicit-assessment
- Fragment
- Be-positive
- Inform
- Be-negative
- Elicit-inform
- Comment-about-understanding
- Suggest
- Elicit-comment-about-understanding
- Offer
- Elicit-offer-or-suggest
- Other

— Dynamic Interpretation Theory (DIT) (Bunt, 1994) reported dialogue acts schema with a number of dialogue act types from DAMSL(Allen and Core, 1997) and other schema. The DIT++ is a comprehensive system of dialogue act types obtained by extending the acts of DIT(Bunt and Girard, 2005). Moreover, DIT++ schema has 11 dimensions with around 95 communicative functions, around 42 of which, like switchboard are for general purpose functions, whereas others cover elements of feedback, interaction management and the control of social obligations(Webb, 2010).

— (Bunt et al., 2010) has proposed a preliminary version of ISO DIS 24617-2:2010 as an international standard for annotating dialogue with semantic information; in particular concerning the communicative functions of the utterances, the kind of content they address, and the dependency relations to what was said and done earlier in the dialogue. (Bunt et al., 2012) has proposed the final version of ISO Standard 24617-2 as:

- **Task**
  - question
  - propositionalQuestion
  - setQuestion
  - checkQuestion
  - choiceQuestion
  - inform
  - agreement
  - disagreement
  - correct
  - answer
  - confirm
  - disconfirm
  - promise
  - offer
  - addressRequest
  - acceptRequest
  - declineRequest
  - addressSuggest
  - acceptSuggest
  - declineSuggest
  - instruct
  - addressOffer
  - suggest
  - acceptOffer
  - declineOffer
- **Feedback**
  - autoPositive
  - alloPositive
  - autoNegative
  - alloNegative
  - feedbackElicitation
- **Turn Management**
  - turnAccept
  - turnAssgin
  - turnGrab
  - turnKeep
  - turnRelease
  - turnTake
- **Time Management**
  - stalling
  - pausing
- **Own & Partner Communication Management**
  - complete
  - correctMisspeaking
  - selfError
  - retraction
  - selfCorrection
- **Discourse Structuring**
  - interactionStructuring
  - opening
- **Social Obligation management**
  - initialGreating
  - returnGreating
  - initialSelfIntroduction
  - returnSelfIntroduction
  - apology
  - acceptApology
  - thanking
  - acceptThanking
  - initialGoodbye
  - returnGoodbye

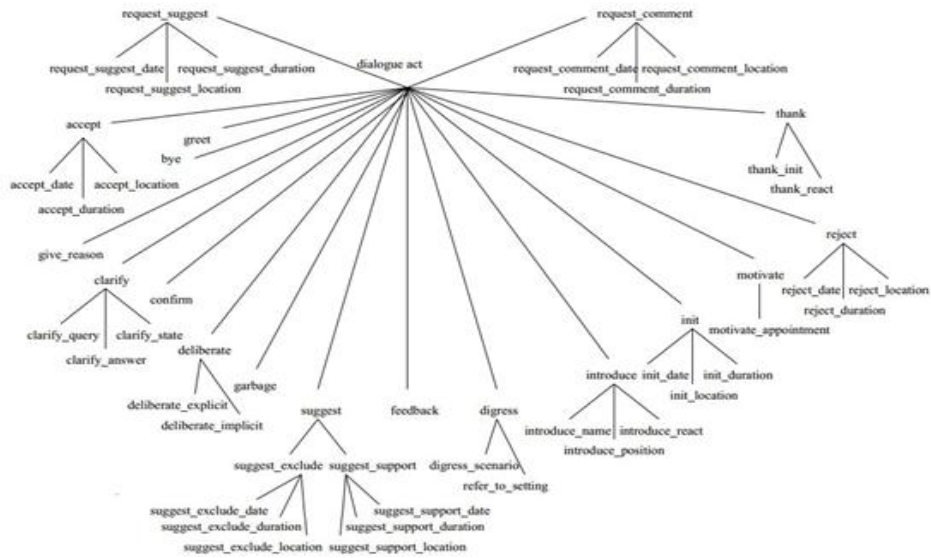

Figure 2: VERBMOBIL project 1st phase hierarchy of 43 dialogue acts (Jekat et al., 1995)

As the best of our knowledge, all of the previous dialogue acts annotation schemas applied to mark-up dialogue corpora based on non-Arabic languages such as English, German and Spanish.

Moreover, there are few efforts were done to propose dialogue acts annotation schemas for Arabic such as

- So, the first attempt was by (Shala et al., 2010) that proposed dialogue acts schema contains 10 DAs:

    - Assertion
    - Response to Question
    - Command
    - Short Response
    - Declaration
    - Greetings
    - Promise/Denial
    - Expressive Evaluation
    - Question
    - Indirect
    - Request

- (Dbabis et al., 2012) has been improved (Shala et al., 2010) schema; the reported schema based on multi-dimension "6$^{th}$ categories" 13 DAs:

    - **Social Obligation Management**
        - Opening
        - Closing
        - Greeting
        - Polite Formula
        - Introduce
        - Thanking
        - Apology
        - Regret
    - **Turn Management**
        - Acknowledgement
        - Calm
        - Clarify
    - Clarify
    - Feedback
    - Out of topic
    - Non understanding signal
    - **Request**
        - Question
        - Order
        - Promise
        - Hope
        - Wish
        - Invocation
        - Warning
    - **Argumentation**
        - Opinion
    - Appreciation
    - Disapproval
    - Accept
    - Conclusion
    - Partial Accept Reject
    - Partial Reject
    - Argument
    - Justification
    - Explanation
    - Confirmation
    - **Answer**
    - **Statement**

These schemas have applied to mark-up dialogues corpora based on a general conversion discussion like TV talk-show programs.

- (Graja et al., 2013) reported a words semantic labelling schema to mark-up dialogue utterance word-by-word for inquiry-answer dialogues specially train railway stations; this schema contains about 33 semantic labels for word annotation within five dimensions:

    - **Domain concepts**
        - Train
        - Train_Type
        - Departure_hour
        - Arrival_hour
        - Day
        - Origin
        - Destination
        - Fare
        - Class
        - Ticket_Numbers
        - Ticket
        - Hour_Cpt
    - Departure_Cpt
    - Arrival_Cpt
    - Price_Cpt
    - Class_Cpt
    - Trip_time
    - Ticket_type
    - **Requests concepts**
        - Path_Req
        - Hour_Req
        - Booking_Req
        - Price_Req
        - Existence_Req
        - Trip_timeReq
    - Clarification_Req
    - **Dialogue concepts**
        - Rejection
        - Acceptance
        - Politeness
        - Salutation (Begin)
        - Salutation (End)
    - **Link concepts**
        - Choice
        - Coordination
    - **Out of vocabulary**
        - Out

— Recently, (Elmadany et al., 2014) reported a schema based request and response dimensions for inquiry-answer dialogues such as flights, mobile service operators, and banks; this schema contains DAs:

- o Request Acts
  - Taking-Request
  - Service-Question
  - Confirm-Question
  - YesNo-Question
  - Choice-Question
  - Other-Question
  - Turn-Assign
- o Response Acts
  - Service-Answer
  - Other-Answer
  - Agree
  - Disagree
  - Greeting
  - Inform
  - Thanking
  - Apology
  - MissUnderstandingSign
  - Correct
  - Pausing
  - Suggest
  - Promise
  - Warning
  - Offer
- o Other Acts
  - Opening
  - Closing
  - Self-Introduce

## 3.2 Arabic Dialogue Acts Corpora

The use of corpora has been a key player in the recent advance in NLP research. However, the high costs of licensing corpora could be a difficult for many young researchers. Therefore, find freely available corpora is clearly a desirable goal, unfortunately; the freely available corpora are mostly not easily found and the most resources available from language data providers are expenses paid or exclusively reserved for subscribers. As the best of our knowledge, Arabic dialogue segmentation processing is considered hard due to the special nature of the Arabic language and the lake of Arabic dialogues segmentation corpora (Zaghouani, 2014). However, there are many annotated dialogued acts corpora for non-Arabic languages, these are the most annotated corpora used in DAs classifications tasks listed in(Webb, 2010) for non-Arabic languages such as:

— **MAPTASK**[1]: consist of 128 English dialogues, containing 150,000 words.
— **VERBMOBIL**[2]: consist of 168 English dialogues, containing 3117 utterances. This corpus has annotated with 43 distinct Dialogue Acts.
— **SWITCHBOARD**[3]: consist of 1155 telephone conversations, containing 205,000 utterances and 1.4 million words.
— **AMITIES**[4]: consist of 1000 English human-human dialogues from GE call centres in the United Kingdom. These dialogues containing 24,000 utterances and a vocabulary size of around 8,000 words.
— **AMI**[5]: Contains 100 hours of meeting.

Unfortunately, to found fully Annotated Arabic dialogue acts corpus is more difficult but there are many of Arabic speech corpora prepared for Automatic Speech Recognition (ASR) research/application. Moreover, most of these corpora are available from the LDC or ELRA members with membership fees e.g. CALLHOME corpus [6](Canavan et al., 1997). Therefore, as the best of our knowledge, there are some efforts to building a fully annotated corpus for Arabic dialogues such as:

— TuDiCoI[6] (Tunisian Dialect Corpus Interlocutor): Corpus consists of Railway Information from the National Company of Railway in Tunisia (SNCFT) which a transcribed spoken Arabic dialogues; these dialogues are between the SNCFT staff and clients who request information about the train time, price, booking...etc. Moreover, the initial corpus of TuDiCoI has reported by (Graja et al., 2010) containing 434 transcribed dialogues with 3080 utterances includes 1465 staff utterances and 1615 client utterances. So, TuDiCoI corpus has enriched by(Graja et al., 2013) to contain 1825 transcribed dialogues with 12182 utterances includes 5649 staff utterances and 6533 client utterances. In addition, each dialogue consist of three utterances for clients and three utterances for staff; client turn is composed of average 3.3 words. The low words per clients utterances and dialogues length is due to the words used by clients to request for information about railway services. Moreover, the corpus turns are not segmented into utterances because it is sort and they considered the utterance is equal to the turn as shown in Table 1. Unfortunately, TuDiCoI are not annotated using DAs schema but it is marked-up by word-by-word schema (see section 3.1) as shown in Figure 3.

---

[1] Available at http://www.hcrc.ed.ac.uk/maptask/
[2] Available at http://www.phonetik.uni-muenchen.de/Bas/Bas Korporaeng.html
[3] Available at ftp://ftp.ldc.upenn.edu/pub/ldc/public-data/swb1_-dialogact-annot.tar.gz
[4] Available at http://www.dcs.shef.ac.uk/nlp/amities/
[5] Available at http://groups.inf.ed.ac.uk/ami/corpus/

Table 1. A sample of TuDiCoI real dialogue (Graja et al., 2013)

| Persons | Utterance ID | | Utterances |
|---|---|---|---|
| Customer | U1 | Arabic: | لتونس التران يخرج وقتاش سامحني |
| | | *Buckwalter:* | *sAmHny wqtA$ yxrj EttrAn ltwns* |
| | | English: | Excuse me when the train leaves to Tunis |
| Operator | U2 | Arabic: | أدراج وربعه ساعه ماضي |
| | | *Buckwalter:* | *mADy sAEh wrbEh OdrAj* |
| | | English: | One hour past twenty minutes |
| Customer | U3 | Arabic: | التكيه هوا بقداش |
| | | *Buckwalter:* | *bqdA$ hwA Ettkyh* |
| | | English: | How much the ticket |
| Operator | U4 | Arabic: | لتونس وخمسميه نلف ثناش |
| | | *Buckwalter:* | *vnA$ nlf wxmsmyh ltwns* |
| | | English: | Twelve dinars and five hundred to Tunis |

(Elmadany et al., 2014) is reported a manually annotated Arabic dialogue acts corpus and manually segmented turns into utterances for Arabic dialogues language understanding tasks. It has contains an 83 Arabic dialogues for inquiries-answers domains which are collected from call-centers as shown in

— Table 2Table 2. Moreover, this corpus contains two parts:

o   Spoken dialogues, which contains 52 phone calls recorded from Egyptian's banks and Egypt Air Company call-centers with an average duration of two hours of talking time after removing ads from recorded calls, and It consists of human-human discussions about providing services e.g. Create new bank account, service request, balance check and flight reservation. Moreover, these phone calls have transcribed using Transcriber®[7], a tool that is frequently used for segmenting, labeling and transcribing speech corpora.

---

[6] Available at https://catalog.ldc.upenn.edu/LDC96S35
[7] Available at https://sites.google.com/site/anlprg/outils-et-cor pus-realises/TuDiCoIV1.xml?attredirects=0
[8] http://trans.sourceforge.net/en/presentation.php

o      Written 'Chat' dialogues, which contain 31 chat dialogues, collected from mobile network operator's online-support 'KSA Zain, KSA Mobily, and KSA STC'.

Table 2: Segmented Arabic Dialogue sample from (Elmadany et al., 2014) corpus

| Turn ID | Persons | Utterance ID | Utterances | |
|---|---|---|---|---|
| T1 | Operator | U1 | Arabic:<br>*Buckwalter:*<br>English: | ان اس جي بي<br>*An As jy by*<br>NSBG |
| | | U2 | Arabic:<br>*Buckwalter:*<br>English: | شريفة المصري<br>*$ryfp AlmSry*<br>Sherifa Elmasri |
| | | U3 | Arabic:<br>*Buckwalter*:<br>English: | مساء الخير<br>*msA' Alxyr*<br>Good afternoon |
| T2 | Customer | U4 | Arabic:<br>*Buckwalter*:<br>English: | الو<br>*Alw*<br>Allo |
| | | U5 | Arabic:<br>*Buckwalter:*<br>English: | مساء الخير<br>*msA' Alxyr*<br>Good afternoon |
| T3 | Operator | U6 | Arabic:<br>*Buckwalter:*<br>English: | مساء النور<br>*msA' Alnwr*<br>Good afternoon |
| T4 | Customer | U7 | Arabic:<br>*Buckwalter:*<br>English: | من فضلك<br>*mn fDlk*<br>If you please |
| | | U8 | Arabic:<br>*Buckwalter:*<br>English: | كنت عايزة اسأل عن قروض السيارات<br>*knt EAyzp As>l En qrwD AlsyArAt*<br>I want to ask about cars loan |
| | | U9 | Arabic:<br>*Buckwalter:*<br>English: | بس هو المشكلة انني معنديش اصلا حساب عندكم<br>*bs hw Alm$klp Anny mEndy$ ASlA HsAb Endkm*<br>The problem is I haven't an account in your bank |
| T5 | Operator | U10 | Arabic:<br>*Buckwalter*:<br>English: | اتشرف بالاسم ايه يا فندم<br>*At$rf bAlAsm Ayh yA fndm*<br>Your name please |
| T6 | Customer | U11 | Arabic:<br>*Buckwalter*:<br>English: | عبير<br>*Ebyr*<br>Abeer |
| T7 | Operator | U12 | Arabic:<br>*Buckwalter*:<br>English: | اهلا وسهلا بيكي<br>*AhlA wshlA byky*<br>You are welcome |
| T8 | Customer | U13 | Arabic:<br>*Buckwalter*:<br>English: | اهلا بيك<br>*AhlA byk*<br>Welcome |
| T9 | Operator | U14 | Arabic:<br>*Buckwalter:*<br>English: | استاذة عبير<br>*AstA*p Ebyr*<br>Miss Abeer, |
| | | U15 | Arabic:<br>*Buckwalter:*<br>English: | حضرتك تقدري تكلمينا علي 19084<br>*HDrtk tqdry tklmynA Ely 19084*<br>You can call us in 19084 |
| T10 | Customer | U16 | Arabic:<br>*Buckwalter*: | 19084<br>*19084* |

| | | | English: | 19084 |
|---|---|---|---|---|
| T11 | Operator | U17 | Arabic:<br>*Buckwalter*:<br>English: | اه<br>*Ah*<br>Yes |
| | | U18 | Arabic:<br>*Buckwalter*:<br>English: | ده الخط الساخن لقروض السيارات يا استاذة عبير<br>*dh AlxT AlsAxn lqrwD AlsyArAt yA AstA\*p Ebyr*<br>This is cars loan hotline, Miss Abeer |
| T12 | Customer | U19 | Arabic:<br>*Buckwalter*:<br>English: | اوكيه ماشي<br>*Awkyh mA$y*<br>Ok |
| T13 | Operator | U20 | Arabic:<br>*Buckwalter*:<br>English: | اي استفسار تاني<br>*Ay AstfsAr tAny*<br>Any other service? |
| T14 | Customer | U21 | Arabic:<br>*Buckwalter*:<br>English: | ميرسي<br>*myrsy*<br>No thanks |
| T15 | Operator | U22 | Arabic:<br>*Buckwalter*:<br>English: | شكرا علي اتصال حضرتك<br>*$krA Ely AtSAl HDrtk*<br>Thanks for your calling |

[Departure_Cpt] يمشي [Train] التران [Out] إي [Out] إي [Hour_Req] وقتاش [Out] مع [

Figure 3. TuDiCoI semantic labelling (Graja et al., 2013)

Building an annotated DAs corpus need four process recoding (for spoken)/ collecting (for chat) dialogues process, transcription process (for spoken only), segmentation process, and annotation process. Moreover, these processes are expensive.

### 3.3 Arabic Dialogue Segmentation

A segmentation process generally means dividing the long unit into meaningful pieces or small units "non-overlapping units" and it is considering one of the important solutions to solve Natural Language Processing (NLP) problems. Definition of segmentation will differ according to the NLP problem such as:

1. When dividing the text into topics, paragraphs, or sentences, properly named Text Segmentation e.g. (Touir *et al.*, 2008; El-Shayeb *et al.*, 2007).
2. When dividing the sentences into a group of words, properly named Phrase Segmentation.
3. When dividing words into its clitics/affix (prefix, stem, and suffix), properly named tokenization e.g. (Diab *et al.*, 2004).

Build a completely Human-Computer systems and the belief that will happens has long been a favourite subject in research science. So, dialogue language understanding is growing and considering the important issues today for facilitate the process of dialogue acts classification; consequently segment the long dialogue turn into meaningful units namely utterances is increasing. Moreover, Human-Computer Dialogues are divided into different types: Speech Dialogues proper name "Spoken Dialogue" which works in waves and Written Dialogues proper name "Chat" or "Instant Massaging" (IM) which works on text. The waveform in spoken dialogues is usually segment the long input into short pieces based on simple acoustic criteria namely pauses "non-speech intervals", this type of segmentation is namely acoustic segmentation; but it's different when working in text such as chat dialogues, here use a linguistic segmentation. Consequently, to improve the human-computer system need for understand spoken dialogue by extracting the meaning of speaker's utterances, the acoustic segmentation is inadequate in such cases that are needed for further processing based on syntactically and semantically coherent units because it is not reflecting the linguistic structure of utterances(Stolcke and Shriberg, 1996). However, segmentation process is known in dialogues language understanding by many titles such as Utterances Segmentations, Turns Segmentations, and Dialogue Acts Segmentations (see section 2.2);

There are many approaches to understanding both dialogues types (spoken and written) for non-Arabic languages e.g. English, Germany, France... etc. (Ang et al., 2005; Ivanovic, 2005; Zimmermann et al., 2005; Ding and Zong, 2003). Moreover, understanding Arabic dialogues have gained an increasing interest in the last few years. To the best of our knowledge; there are few works interested in Arabic dialogue acts classification (see section 3.4); these works have used the user's *turn* as an *utterance* without any segmentation e.g. (Shala et al., 2010; Bahou et al., 2008; Graja et al., 2013; Lhioui et al., 2013; Hijjawi et al., 2013; Hijjawi et al., 2014). In addition, there are a few works for the Arabic discourse segmentation such as:

— (Belguith et al., 2005) has proposed a rule-based approach based on 83 rules for Arabic text segmentation which extracted from contextual analysis of the punctuation marks, the coordination conjunctions and a list of particles that are considered as boundaries between sentences.

— (Touir et al., 2008)has proposed a rule-based approach based on sentences connectors without relying on punctuation based on empirical study of 100 Articles, each article have between 450 and 800 words, for analysis to extract the connectors. Consequently, they provided term "*Passive*" for connector that does not imply any cutting point e.g. "و /and /w" and term "*Active*" for connector which indicates the beginning or the end of a segment e.g. "لكن /but /lkn". In addition, they concluded that *Passive* connector has useful only when comes before *active*. Hence, they are tested the approach on 10 articles, each article have 500 to 700 words.

— (Khalifa et al., 2011) proposed a Machine-Learning approach using SVM based on the connector "و/and/w". Moreover, they reported sixth types of "و /and /w" connector that divided into two classes: (1) "*Fasl*" for a connector that indicates the beginning of segments, and (2) "*Wasl*" for connector that does not have any effect on segmentation. In additional, they are built a corpus for newspapers and books which includes 293 instances of the connector "و /and /w" and added diacritization marks manually to the corpus text (training and testing) during the preparation steps. However, these approach very similar to (Touir et al., 2008) when considering the connector "و /and /w".

— (Keskes et al., 2012) proposed a rule-based approach based on three principals: (1) using punctuation indicators principal only (2) using lexical cues principal only (3) using mixed punctuation indicators and lexical cues. In addition, they used 150 news articles (737 paragraphs, 405332 words) and 250 elementary school textbooks (1095 paragraphs, 29473 words) for built the lexical cues and effective punctuation indicators. Moreover, they concluded two types of punctuation indicators: (1) "*strong*" that always identify the end or the start of the segments such as the exclamation mark (!), the question mark (?), the colon (:) and the semi-colon (;) (2) "*Weak*" that don't always identify the begin or the begin of the segment segments such as full-stop (.), the comma (,), quotes, parenthesis, brackets, braces and underscores; They reported the mixed punctuation indicators and lexical cues principal has the best results in textbooks and newspapers.

International Journal on Natural Language Computing (IJNLC) Vol. 4, No.2, April 2015These approaches are not testing on Arabic dialogues that completely differs for newspapers and books articles; and Arabic spontaneous dialogues is properly dialect Arabic, which is informal text.

### 3.4. Arabic Dialogue Acts Classification

There are two ways to understand the dialogues language(Webb and Hardy, 2005):

— **Shallow understanding:** It is simple spotting keywords or having lists of, for example, every location recognized by the system. Several systems are able to decode directly from the acoustic signal into semantic concepts precisely because the speech recognizer already has access to this information.

— **Deeper analysis:** Using linguistic methods; including part-of-speech tagging, syntactic parsing and verb dependency relationships.

Using Machine Learning (ML) for solving the DA classification problem, researchers have not historically published the split of training and testing data used in their experiments, and in some cases methods to reduce the impact of the variations that can be observed when choosing data for training and testing have not been used (Webb, 2010). Moreover, DAs are practically used in many live dialogue systems such as Airline Travel Information Systems (ATIS) (Seneff et al., 1991), DARPA (Pellom et al., 2001), VERBMOBIL project(Wahlster, 2000), and Amities dialogue system(Hardy et al., 2004). Now, we will describe in brief some of DAs approaches over annotated corpora to recognize dialogue acts:

— Several approaches have proposed for DAs classification and N-gram models can be considering the simplest method of DA prediction; predicting the upcoming, DA based on some limited sequence of previous DAs such as(Hardy et al., 2004; Webb, 2010; Webb and Hardy, 2005; Webb et al., 2005a, 2005b; Nagata and Morimoto, 1994; Niedermair, 1992) . Moreover, (Hardy et al., 2004; Webb, 2010; Webb and Hardy, 2005; Webb et al., 2005a, 2005b; Nagata and Morimoto, 1994; Niedermair, 1992)are used Hidden Markova Model (HMM) with N-gram.

— Samuel et al. (1998) used Transformation-Based Learning (TBL) (Brill, 1995)over a number of utterance features, including utterance length, speaker turn and the dialogue act tags of adjacent utterances.

— (Carberry and Lambert, 1999) used a rule-based model of DA recognition that uses three sources of knowledge, linguistic (including cue phrases), contextual and world knowledge. Moreover, the linguistic knowledge is used primarily to identify if the speaker has some belief in the evidence presented, using prior known cue phrases e.g. BUT, or the use of surface-negative question forms (Doesn't X require Y?) (Webb, 2010). Also (Prasad and Walker, 2002) are used a rule based learning method in the DARPA Communicator dialogues. More recently, (Georgila et al., 2009) extended (Prasad and Walker, 2002) work to include manually constructed context rules that cover the user side of the Communicator dialogues

— Bayesian approaches have proven to be effective for DAs classification(Webb, 2010); (Grau et al., 2004)used Naïve Bayesian over the WITCHBOARD corpus within a tri-gram language model.

— (Ji and Bilmes, 2005; Ji and Bilmes, 2006) are investigated the use of dynamic Bayesian networks (DBNs) using graphical models and they reported the best performing set of features is a tri-gram model of the words in the utterances combined with a bi-gram model of DA.

These approaches are tested on non-Arabic dialogues e.g. English, Germany, France... etc. which completely differs for Arabic dialogues. Moreover, understanding Arabic dialogues have gained an increasing interest in the last few years. To the best of our knowledge, there are few works interested in Arabic dialogue acts classification such as:

— (Bahou et al., 2008) proposed a method for the semantic representations of utterances of spontaneous Arabic speech based on the frame grammar formalism as show in Figure 4 and it's tested on Tunisian national railway queries (1003 queries representing 12321 words) collected using Wizard-of-Oz technology. In addition, this method consists of three major steps: a pre-treatment step that includes the normalization of the utterance and its morphological analysis; a step of semantic analysis that assigns semantic tags to each lexical unit of query; and a frame generation step that identifies and fills the semantic frames of the utterance. They reported 37% recall, 60.62% precision and 71.79% as F-Measure for classification with average execution time for the utterance is 0.279 sec.

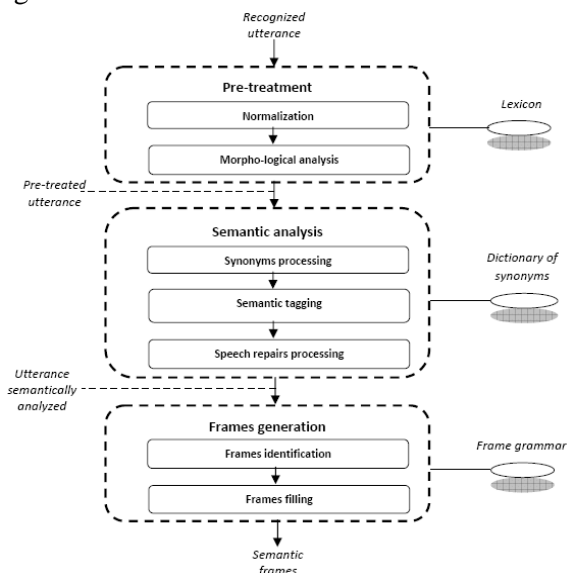

Figure 4. (Bahou et al., 2008) Approach

— (Shala et al., 2010) proposed a fully automated method for speech act classification for Arabic discourse based on the hypothesis that the initial words in a sentence and/or their parts-of-speech are diagnostic of the particular speech act expressed in the sentence. In addition, used the semantic categorization of these words in terms of named entities and combined this approach with Support Vector Machines (SVM) models to automatically derive the parameters of the models they used to implement the approach as show in Figure 5. Moreover, they used two machine-learning algorithms, Naïve Bayes and Decision Trees to induce classifiers acts for Arabic texts and they reported 41.73% as accuracy scores of all models.

International Journal on Natural Language Computing (IJNLC) Vol. 4, No.2,April 2015

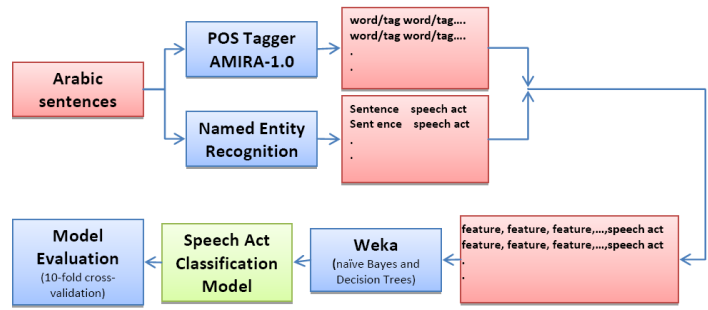

Figure5. (Shala et al., 2010) Approach

— (Lhioui et al., 2013) proposed an approach based on syntactic parser for the proper treatment of utterances including certain phenomena such as ellipses and it has relies on the use of rule-base (context free grammar augmented with probabilities associated with rules) as show in Figure 6. In addition, they used HHM for creating the stochastic model (if a pretreated and transcribed sequence of words - this words are obviously the output of recognition module - and their annotated corresponding sequences was taken). Moreover, they applied their method on Tunisian touristic domain collected using Wizard-of-Oz technology which contains 140 utterances recorded from 10 speakers with 14 query types (DA) e.g. negation, affirmation, interrogation and acceptance and reported 70% recall, 71% precision and 73.79% as F-measure for classification with average execution time 0.29 seconds to process an utterance of 12 words

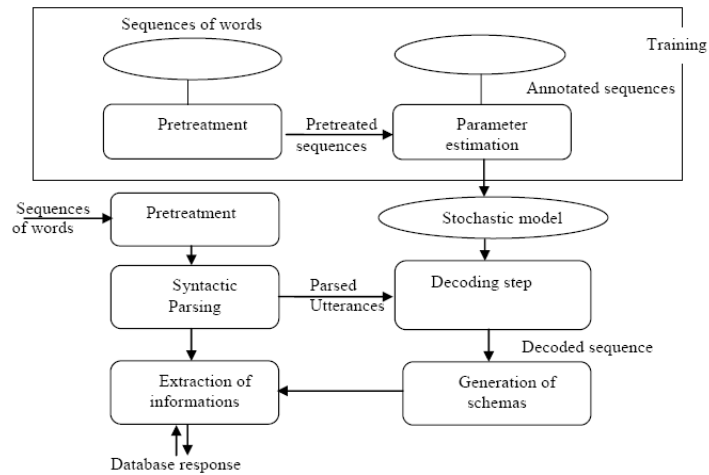

Figure 6. (Lhioui et al., 2013) Approach

— (Graja et al., 2013) proposed discriminative algorithm based on Conditional Random Fields (CRF)[8] to semantically label spoken Tunisian dialect turns which are not segmented into utterances from TuDiCoI corpus (see section 3.2) as show in Figure 3. Moreover, they applied some treatments to improve turn's structure: (1) lexical normalization such as replacing the word "رزرفسيون" "Reservation" for all its forms e.g. "رزرفسيون", "رازرفسيون", "رازارفسيون", "ريزرفسيون". (2) Morphological analysis and lemmatization such as replacing the word "خارج" "*is going*" and "يخرج" "*goes*" by the following canonical form "خرج" "*go*". (3) Synonyms treatment, this treatment consists in replacing each word by its synonym. In addition, they applied the approach on two data sets one without the treatments and the second with the treatments; and they reported that the treatments has reduce the errors rate compared to the non-treatments data set from 12% to 11%.

---

[9] Conditional random fields (CRF) are undirected graphical models trained to maximize a conditional probability which proposed by (Lafferty et al., 2001)

— (Hijjawi et al., 2013) proposed approach based on Arabic function words such as "هل" "do/does", "كيف" "How" and it's focused on classifying questions and non-questions utterances. Moreover, the proposed approach extracts function words features by replacing them with numeric tokens and replacing each content word with a standard numeric token; they used the Decision Tree to extract the classification rules and this approach used on Conversational Agent called ArabChat (Hijjawi et al., 2014) to improve its performance by differentiating among question-based and non-question-based utterances.

— (Neifar et al., 2014) update (Bahou et al., 2008) approach to understanding Tunisian dialect using lexical database and conceptual segmentation. They used TuDiCoI corpus in evaluation.

## 4.CONCLUSIONS

We presented this survey for the Arabic dialogues language understanding or Arabic dialogue Acts classification and the goal behind this study is to promote the development and use of Human-Computer research in Arabic dialogues. The results obtained showed that a few works that developed based on Arabic dialogues. Consequently, we hope that this initial attempt to increasing and improve this research as non-Arabic languages.